\documentclass[letterpaper]{article} 
\usepackage{aaai24}  
\usepackage{times}  
\usepackage{helvet}  
\usepackage{courier}  
\usepackage[hyphens]{url}  
\usepackage{graphicx} 
\usepackage{natbib}  
\usepackage{caption} 
\usepackage{algorithm}

\usepackage{newfloat}
\usepackage{listings}
\DeclareCaptionStyle{ruled}{labelfont=normalfont,labelsep=colon,strut=off} 
\floatstyle{ruled}
\newfloat{listing}{tb}{lst}{}
\floatname{listing}{Listing}
\usepackage{amsfonts,bm}
\usepackage{url}
\usepackage{nicefrac}       
\usepackage{colortbl}
\usepackage{enumitem}
\usepackage{adjustbox}
\usepackage{multirow}

\usepackage[table,dvipsnames]{xcolor}
\definecolor{grey}{RGB}{128,138,135}
\definecolor{darkgrey}{RGB}{96,96,96}
\definecolor{lavender}{HTML}{E5E2FB}
\definecolor{lightblue}{HTML}{CEEBF9}
\definecolor{lightyellow}{HTML}{F7D9AE}
\definecolor{lightred}{HTML}{EEDCDB}
\definecolor{green}{HTML}{3BCB41}
\definecolor{darkgreen}{HTML}{156C09}
\definecolor{purple}{HTML}{9903F0}

\def\myEqref#1{Eq.~(\ref{#1})}
\def\vtheta{{\bm{\theta}}}
\def\va{{\bm{a}}}
\def\vb{{\bm{b}}}
\def\vg{{\bm{g}}}
\def\vx{{\bm{x}}}
\def\vy{{\bm{y}}}

\def\mA{{\bm{A}}}
\def\mW{{\bm{W}}}
\def\mX{{\bm{X}}}
\def\mY{{\bm{Y}}}
\def\gO{{\mathcal{O}}}
\newcommand{\E}{\mathbb{E}}
\newcommand{\Ls}{\mathcal{L}}

\usepackage{mathtools}

\DeclareMathOperator*{\argmin}{arg\,min}

\usepackage[capitalize]{cleveref}
\crefname{section}{Sec.}{Secs.}
\Crefname{section}{Section}{Sections}
\Crefname{table}{Table}{Tables}
\crefname{table}{Tab.}{Tabs.}

\usepackage{empheq}

\usepackage{booktabs,arydshln}
\makeatletter
\def\adl@drawiv#1#2#3{%
        \hskip.5\tabcolsep
        \xleaders#3{#2.5\@tempdimb #1{1}#2.5\@tempdimb}%
                #2\z@ plus1fil minus1fil\relax
        \hskip.5\tabcolsep}
\newcommand{\cdashlinelr}[1]{%
  \noalign{\vskip\aboverulesep
           \global\let\@dashdrawstore\adl@draw
           \global\let\adl@draw\adl@drawiv}
  \cdashline{#1}
  \noalign{\global\let\adl@draw\@dashdrawstore
           \vskip\belowrulesep}}
\makeatother

\DeclareCaptionLabelFormat{figtab}{#1~#2 \& \tablename~\thetable}

\usepackage{amsthm}
\newtheorem*{remark}{Remark}

\usepackage[noend]{algpseudocode}

\usepackage{pifont}
\newcommand{\cmark}{\ding{51}}%
\newcommand{\xmark}{\ding{55}}%

\usepackage{xspace}
\makeatletter
\DeclareRobustCommand\bmvaOneDot{\futurelet\@let@token\bmv@onedotaux}
\def\bmv@onedotaux{\ifx\@let@token.\else.\null\fi\xspace}
\makeatother
\def\eg{\emph{e.g}\bmvaOneDot}

\def\etal{\emph{et al}\bmvaOneDot}

\def\ie{\emph{ie}\bmvaOneDot}

\def\wrt{w.r.t\bmvaOneDot} 

\def\with{\textbf{w/}\xspace}
\def\without{\textbf{w/o}\xspace}

\usepackage{amssymb}

\title{Concealing Sensitive Samples against Gradient Leakage in Federated Learning}
\author{
    Jing Wu\textsuperscript{\rm 1},
    Munawar Hayat\textsuperscript{\rm 2},
    Mingyi Zhou\textsuperscript{\rm 3},
    Mehrtash Harandi\textsuperscript{\rm 1}
}
\affiliations{
    \textsuperscript{\rm 1}Department of Electrical and Computer Systems Engineering, Monash University \\
    \textsuperscript{\rm 2}Department of Data Science and AI, Monash University\\
    \textsuperscript{\rm 3}Department of Software Systems and Cybersecurity, Monash University \\
    \texttt{\{jing.wu1, munawar.hayat, mingyi.zhou, mehrtash.harandi\}@monash.edu} \\
}

\usepackage{bibentry}

\begin{document}

\maketitle

\begin{abstract}
Federated Learning (FL) is a distributed learning paradigm that enhances users' privacy by eliminating the need for clients to share raw, private data with the server.
Despite the success, recent studies expose the vulnerability of FL to model inversion attacks, where adversaries reconstruct users’ private data via eavesdropping on the shared gradient information. 
We hypothesize that a key factor in the success of such attacks is the low entanglement among gradients per data within the batch during stochastic optimization. This creates a vulnerability that an adversary can exploit to reconstruct the sensitive data.
Building upon this insight, we present a simple, yet effective defense strategy that obfuscates the gradients of the sensitive data with concealed samples. To achieve this, we propose synthesizing concealed samples to mimic the sensitive data at the gradient level while ensuring their visual dissimilarity from the actual sensitive data.
Compared to the previous art, our empirical evaluations suggest that the proposed technique provides the strongest protection while simultaneously maintaining the FL performance.
Code is located at \url{https://github.com/JingWu321/DCS-2}.
\end{abstract}

\section{Introduction}
\label{sec: intro}
Consider an Artificial Intelligence (AI) service that aids in disease diagnosis. Multiple hospitals train a model for this service in collaboration. Publishing such a service could benefit a large number of doctors and patients, but it is critical to ensure that private medical data is secure and the utility of the service is normal.
Federated Learning (FL)~\cite{mcmahan2017communication} is an essential technology for such critical applications where the confidentiality of private data is important.
FL provides a distributed learning paradigm that enables multiple clients (\eg, hospitals, businesses, or even mobile devices) to train a unified model jointly under the orchestration of a central server. 
A key advantage of FL lies in its promise of privacy for participating clients. With data decentralized and users' information kept solely with the client, only model updates (\eg, gradients) are transmitted to the central server.
Since the model's updates are specifically tailored to the learning task, they may create a false sense of security for FL clients, leading them to believe that the shared updates contain no information on their private training data~\cite{kairouz2021advances}.

Recent \textit{model inversion attacks}~\cite{zhu2019deep, geiping2020inverting, balunovic2021bayesian,fowl2021robbing, li2022auditing} have shown that the users' private data can be reconstructed from the gradients shared during the learning process. This alarming finding has led to the exploration of various defense schemes to mitigate privacy leakage. Zhu \etal~\cite{zhu2019deep} employed a strategy that adds noise to gradients, guided by Differential Privacy (DP)~\cite{dwork2006calibrating, abadi2016deep, song2013stochastic, mcmahan2017learning}, a concept originally designed to constrain information disclosure.
They also utilized gradient compression~\cite{lin2017deep}, which prunes gradients below a threshold magnitude, as a protective measure.
Latest techniques have further advanced the field, with developments such as Automatic Transformation Search (ATS)~\cite{gao2021privacy} (augmenting data to hide sensitive information), PRivacy EnhanCing mODulE (PRECODE)~\cite{scheliga2022precode} (use of bottleneck to hide the sensitive data), and Soteria~\cite{sun2021soteria} (pruning gradients in a single layer).

However, as defense techniques improve, \textbf{attacks evolve} as well. 
New findings, as highlighted by \citet{balunovic2021bayesian} and \citet{li2022auditing}, indicate that modern defenses may be ineffective against more sophisticated attacks.
For example, \citet{balunovic2021bayesian} show that an adversary can disregard the gradients pruned by Soteria and still reconstruct inputs, even without knowledge of the specific layers where pruning is applied. The vulnerability also extends to other defenses; data can be readily reconstructed in the initial communication rounds against the defense ATS~\cite{balunovic2021bayesian}. In the case of the defense PRECODE, the mere presence of a single non-zero entry in the bias term can enable perfect reconstruction by adversaries~\cite{balunovic2021bayesian}.

Most current defenses seek to protect all data equally, even if this results in a poor privacy-performance trade-off.
In this work, we argue for a more realistic and practical setup where the focus should be given to the sensitive data (\eg, personal data revealing racial or ethnic origin, political opinions, and religious beliefs as mentioned in European Union’s General Data Protection Regulation~\cite{voigt2017eu}).
Consider a malignant skin lesion recognition system as an example. Skin images with tattoos that contain personal information demand extra attention than images without such information. As such, preserving the former's privacy should be the algorithms' priority.

Exploring the underlying mechanism of model inversion attacks, we hypothesize that these attacks capitalize on the characteristic of relatively low entanglement among the gradients of data points during stochastic optimization. Building upon this understanding, we introduce a defense strategy that obfuscates the gradients of sensitive data using concealed samples.
Formally, our goal is to ensure that an adversary is unable to reconstruct sensitive data while simultaneously preserving the performance of the FL system. To achieve this, we propose an algorithm that can adaptively synthesize concealed samples in lieu of sensitive data.
We design the concealed points to have high gradient similarity with the sensitive data but visually disparate. For this purpose, our proposed defense has two main characteristics;
\textbf{1) \textbf{Enhancing the privacy of sensitive data.}} Even though the gradients from the concealed data are similar to those of the sensitive data, inverting these gradients results in data points that are visually very different from the sensitive data. 
By obfuscating the gradients of the sensitive data with those of the concealed data, the reconstruction of sensitive information becomes confounded, which in turn leads to enhancing the privacy of sensitive data in FL.
\textbf{2) Maintaining the FL performance.} The introduction of concealed data could potentially disrupt the learning process as it alters the gradient information. Our algorithm mitigates this by ensuring that the shared gradients, after the introduction of concealed data, align with the gradients of the original training samples, including sensitive data. This alignment is achieved through a gradient projection-based approach, preserving the learning capability of the FL system.
Unlike existing defenses, our approach proposes a practical solution to enhance privacy in FL. It presents a significant challenge for an adversary to reconstruct the user-defined sensitive samples, all without sacrificing the overall performance of the FL system.

Our main contributions can be summarized as follows:
\begin{itemize}
    \item We show that model inversion attacks predominantly exploit the characteristic of relatively low entanglement among gradients of samples during stochastic optimization. Based on this finding, we propose to adaptively synthesize concealed samples that obfuscate the gradients of sensitive data.
    \item The proposed approach crafts concealed samples that are adaptively learned to enhance privacy for sensitive data while simultaneously avoiding performance degradation.
    \item We thoroughly evaluate and compare our algorithm against various baselines (\eg, injecting noise to the gradients as in the previous works~\cite{sun2021soteria,gao2021privacy,zhu2019deep}), and empirically observe that our algorithm consistently outperforms the current state-of-the-art defense methods.
\end{itemize}

\section{Related work}
\label{sec: background}

\paragraph{Model Inversion Attacks.}
Several model inversion attacks breach FL privacy by reconstructing the clients' data \eg,~\cite{zhu2020r, fan2020rethinking, zhu2019deep, yin2021see, jin2021catastrophic, jeon2021gradient, li2022auditing, takahashi2023breaching, nguyen2023re}.
Deep Leakage from Gradients (DLG)~\cite{zhu2019deep} and its variants~\cite{zhao2020idlg} employ an optimization-based technique to reconstruct private data from the given gradient updates.
While the original algorithm \cite{zhu2019deep} works best if the number of training samples in each batch is small, subsequent works~\cite{geiping2020inverting, wei2020framework, mo2021quantifying, jeon2021gradient, yin2021see} including Gradient Similarity (GS)~\cite{geiping2020inverting} and GradInversion attack~\cite{yin2021see} are able to reconstruct high-resolution images with larger batch sizes by incorporating stronger image priors.
\citet{jin2021catastrophic} introduce catastrophic data leakage (CAFE) in vertical federated learning (VFL), showing improved data recovery quality in VFL.
\citet{balunovic2021bayesian} formalize the gradient leakage problem within the Bayesian framework and demonstrate that the existing optimization-based attacks could be approximated as the optimal adversary with different assumptions on the input and gradients (\ie, the prior knowledge about the input and conditional probability of the gradient given the input).
They further show that most existing defenses are not quite effective against stronger attacks once appropriate priors (\eg, using generative adversarial networks~\cite{li2022auditing}) are incorporated to reconstruct data.

While aforementioned optimization-based model inversion attacks assume the server is honest-but-curious~\cite{goldreich2009foundations}, recent works \cite{fowl2021robbing, boenisch2021curious} introduce model modification attacks by a malicious server.
\citet{boenisch2021curious} apply trap weights to initialize the model with the goal of activating parts of its parameters, enabling perfect reconstruction within milliseconds.
Similarly, \citet{fowl2021robbing} proposes the insertion of a tailored imprint module into the network structure. The imprinting module will store information exclusively about a specific subset of data points during the updates, and as a result, data can be recovered precisely and quickly, even when aggregated over large batches.

\paragraph{Privacy Preserving defenses.}
Several approaches propose defense against model inversion attacks that breach users' privacy in FL.
We can broadly categorize the existing defenses against model inversion attacks into four categories: gradient compression~\cite{lin2017deep, sun2021soteria} and perturbation~\cite{dwork2006calibrating, abadi2016deep, song2013stochastic}, data encryption~\cite{gao2021privacy, huang2020instahide}, architectural modifications~\cite{scheliga2022precode}, and secure aggregation via changing the communication and training protocol~\cite{bonawitz2017practical, mohassel2017secureml, lee2021digestive, wei2021gradient} (not considered here).
\citet{zhu2019deep} show that gradient compression can help, while \citet{sun2021soteria} propose Soteria, suggesting gradient pruning in a single layer as a defense strategy.
\citet{zhu2019deep} also explore adding Gaussian or Laplacian noise guided by DP~\cite{dwork2006calibrating, abadi2016deep, song2013stochastic, mcmahan2017learning} to prevent data being reconstructed.
ATS relies on heavy data augmentation on training images to hide sensitive information, while InstaHide~\cite{huang2020instahide, huang2021evaluating} encrypts the private data with data from public datasets.
\citet{scheliga2022precode} introduce PRECODE, which inserts a bottleneck to hide the users' data. 
Despite these significant efforts to develop defense schemes against FL attacks, recent works highlight the vulnerabilities of existing defenses. For example, several studies show that DP requires a large number of participants in the training process to converge~\cite{zhu2019deep,gao2021privacy,sun2021soteria}. \citet{balunovic2021bayesian} show that an adversary can get an almost perfect reconstruction after dropping the gradients pruned by Soteria. Balunovi{\'c} \etal~\cite{balunovic2021bayesian} also suggests that it is easy to reconstruct the data using the GS attack in the initial communication rounds against ATS, while \citet{carlini2020private} shows that the private data can be recovered from the encodings of InstaHide~\cite{huang2020instahide, huang2021evaluating}. For PRECODE, \citet{balunovic2021bayesian} demonstrate that an adversary can completely reconstruct the data with at least one non-zero entry in the bias.
Further, strong defenses like Soteria can still be bypassed by the Generative Gradient Leakage (GGL) attack method \cite{li2022auditing}.

\section{Methodology}
\label{sec: alg}
In this section, we outline our proposed defense against model inversion attacks. We begin by introducing a basic FL framework, followed by an explanation of a simple reconstruction formulation that illustrates how model inversion attacks operate with shared gradient information. Subsequently, we describe how our proposed approach counters these attacks. Throughout the paper, we denote scalars by lowercase symbols, vectors by bold lowercase symbols, and matrices by bold uppercase symbols  (\eg, $a$, $\va$, and $\mA$).

\subsection{Federated learning}
Let $f_\vtheta: \mathcal{X} \to \mathcal{Y}$ be a model with parameters $\vtheta$, classifying inputs $\vx \in \mathcal{X}$ 
to labels $\vy$ in the label space $\mathcal{Y}$.
In FL, we assume that there are $C$ clients and a central server. The data $\mathcal{D}_c$ resides with the client $c$, and the server receives the gradient updates from the clients to update the model parameters $\vtheta$ as
\begin{align}
\label{eq: fl}
    \min_{\vtheta} \E_{(\mX, \mY) \sim \mathcal{D}_c} [\Ls (f_\vtheta(\mX), \mY; \vtheta)].
\end{align}

In the $t$-th training round, each client $c$ will compute the gradients $\nabla_\vtheta \Ls (f_\vtheta(\mX), \mY)$ over local training data and send it to the server. The server then updates the model parameters $\vtheta^t$ using gradients from the selected $\tilde{C}$ clients:
\begin{align}
\label{eq: gd}
    \vtheta^t = \vtheta^{t-1} - \frac{\eta}{\tilde{C}} \sum_{c=1}^{\tilde{C}} \nabla_{\vtheta^{t-1}} \Ls (f_\vtheta(\mX), \mY; \vtheta^{t-1}),
\end{align}
where $\eta$ is the learning rate. The server propagates back the updated parameters $\vtheta^t$ to each client, repeating the process until convergence. Even though the private training data never leaves the local clients, in the following, we show how an adversary can still reconstruct the data based on the shared gradients $\nabla_\vtheta \Ls (f_\vtheta(\mX), \mY)$ from client $c$ in the $t$-th communication round.

\begin{remark}
If we assume that each client has its own objective, the FL problem can be formulated as 
\begin{align*}
    \min_{\vtheta} \E_{(\mX, \mY) \sim \mathcal{D}_c} [\Ls_c (f_\vtheta(\mX), \mY; \vtheta)].
\end{align*}
Our solution is generic and can also be used to address this scenario.
\end{remark}

\subsection{Privacy Leakage}

\paragraph{Individual data point leakage.}
Without loss of generality, we consider the case of a network having only one fully connected layer, for which the forward pass is given by $\mathbb{R}^m \ni \vy=\mW^\top \vx +\vb$, where $\mW \in \mathbb{R}^{n \times m}$ is the weight and $\vb \in \mathbb{R}^{m}$ is the bias. Let $\Ls$ denote the objective to update the parameters, then the adversary reconstructs the input $\vx \in \mathbb{R}^{n}$ by computing the gradients of the objective \wrt the weight and the bias:
\begin{align}
\label{eq: signle_grad}
    \nabla_{\mW}\Ls &= [\frac{\partial \Ls}{\partial y_{1}}\frac{\partial y_{1}}{\partial \mW_{:1}}, \cdots, \frac{\partial \Ls}{\partial y_{m}}\frac{\partial y_{m}}{\partial \mW_{:m}}], \notag \\
    \nabla_{\vb}\Ls &= [\frac{\partial \Ls}{\partial y_{1}}, \cdots, \frac{\partial \Ls}{\partial y_{m}}].
\end{align}
Note that $\frac{\partial y_{l}}{\partial \mW_{:l}} = \vx$ for $1 \leq l \leq m$. Thus, we can perfectly reconstruct the input from the gradient information as
$\vx^*=\nicefrac{\nabla_{\mW_{:l}}\Ls}{\nabla_{\vb_{l}}\Ls}=(\frac{\partial \Ls}{\partial y_{l}}\frac{\partial y_{l}}{\partial \mW_{:l}}) / \frac{\partial \Ls}{\partial y_{l}}=\vx$, provided that at least one element of the gradient of the loss with respect to the bias is non-zero (\ie, \(\frac{\partial \mathcal{L}}{\partial y_{l}} \neq 0\), \(1 \leq l \leq m\)).

\paragraph{Multiple data points leakage.}
Let $\vx_j,~ j \in [1, B], B>1$ denotes samples of a mini-batch of size $B$. The gradient of the mini-batch is:
\begin{align}
\label{eq: mp_grad}
    \nabla_{\mW}\Ls &= \frac{1}{B} \sum_{j=1}^B [\frac{\partial \Ls}{\partial y_{1, j}}\frac{\partial y_{1, j}}{\partial \mW_{:1}}, \cdots, \frac{\partial \Ls}{\partial y_{m, j}}\frac{\partial y_{m, j}}{\partial \mW_{:m}}], \notag \\
    \nabla_{\vb}\Ls &= \frac{1}{B} \sum_{j=1}^B [\frac{\partial \Ls}{\partial y_{1, j}}, \cdots, \frac{\partial \Ls}{\partial y_{m, j}}],
\end{align}
which encapsulates a linear combination of all data points $\vx_j$ in the mini-batch.
\citet{sun2021soteria} observe that for data coming from different classes, the corresponding data representations tend to be embedded in different rows/columns of gradients.
Suppose that within the mini-batch, only $\vx_1$ belongs to class $y_c~(1 \leq c \leq m)$, then the column $c$ of the gradient in \cref{eq: mp_grad} will have
\begin{align}
\label{eq: per_grad}
    \frac{\sum_{j=1}^B \frac{\partial \Ls}{\partial y_{c, j}}\frac{\partial y_{c, j}}{\partial \mW_{:c}}}{\sum_{j=1}^B \frac{\partial \Ls}{\partial y_{c, j}}}
    \approx 
    \frac{\frac{\partial \Ls}{\partial y_{c, 1}}\frac{\partial y_{c, 1}}{\partial \mW_{:c}}}{\frac{\partial \Ls}{\partial y_{c, 1}}}
    = \vx_1.
\end{align}
Due to this property, \ie relatively low entanglement among gradients per data points within a batch, the adversary can reconstruct the data in practice.

Boenisch \etal~\cite{boenisch2021curious} also observe that for a ReLU network, over-parameterization can cause all but one training data in a mini-batch to have zero gradients, allowing the individual data point leakage in the mini-batch and the passive adversaries to obtain perfect reconstruction in various cases.

Optimization-based attacks aim to reconstruct data by minimizing the distance between the gradient of the input and that of the reconstruction. In contrast,
model modification attacks utilize specific parameters with the goal of amplifying the leakage of individual data points~\cite{boenisch2021curious} within the mini-batch or allowing portions of the gradient to contain information exclusive to a subset of data points~\cite{fowl2021robbing}.
It is important to note that neither optimization-based attacks nor model modification attacks can precisely separate the gradient for individual data points. 
\textbf{This limitation in the attack algorithms is a vulnerability that we leverage in our approach to protect the data.}

\subsection{Defense by Concealing Sensitive Samples ($\text{DCS}^2$)}
Our objective is to protect sensitive data without modifying any FL settings (\eg, model structure) and the sensitive data themselves, while minimizing the impact of the proposed defense on the model performance.
Previously, we discussed that model inversion attacks reconstruct the inputs using the gradient information since the gradient encapsulates sufficient information about data samples to reconstruct them (see \cref{eq: mp_grad}). We note that while theoretically, attacks cannot precisely separate the gradient for each sample, they can be extremely successful in practice. Our key insight is to insert samples (referred to as concealed samples) to imitate the sensitive data on the gradient level while ensuring that these samples are visually dissimilar to the sensitive data.
Our goal is to make it difficult or even impossible for the adversary to distinguish the gradient of the synthesized concealed samples from the gradient of the sensitive data.

Without loss of generality, assume that there is only one sensitive data point, denoted by \(\vx_{s}\). 
Our task is to construct the concealed sample $\tilde{\vx}_c$ for this sensitive data to achieve the following goals as part of our defense strategy:
\begin{description}[leftmargin=1.0cm,labelindent=0.1cm]
    \vspace{-0.2em}
    \item[Goal-1:] To protect sensitive data from model inversion attacks, we would like to maximize the dissimilarity between the concealed sample $\tilde{\vx}_c$ and the sensitive sample $\vx_{s}$, as measured by $\left\| \tilde{\vx}_c - \vx_{s} \right\|$. Simultaneously, we seek to minimize the similarity between the gradient of the concealed sample \wrt sensitive data. This is quantified by the cosine similarity between the gradient vectors, \ie,
    $\nabla_{\vtheta}\Ls(f_{\vtheta}(\tilde{\vx}_c), \tilde{\vy}_c)$ and $\nabla_{\vtheta}\Ls(f_{\vtheta}(\vx_{s}), \vy_{s})$,
    while ensuring that the resulting latent representation is similar to the sensitive latent representation, \ie, $\left\| f_{\vtheta}(\tilde{\vx}_c) - f_{\vtheta}(\vx_{s}) \right\| \leq \epsilon$.
    \vspace{-0.2em}
    \item[Goal-2:] To facilitate the server's ability to learn and enhance the FL model, we must ensure that the resulting gradient closely resembles the gradient of the batch without concealed samples.
    This can be achieved by satisfying $\langle \nabla_{\vtheta}\Ls(f_{\vtheta}(\{\vx_s\} \cup \{\tilde{\vx}_c\}), \{\vy_{s}\} \cup \{\tilde{\vy}_c\}), \nabla_{\vtheta}\Ls(f_{\vtheta}(\vx_s), \vy_{s}) \rangle > 0$.
\end{description}
To accomplish the aforementioned goals, our defense strategy consists of two phases: \textbf{1.} \emph{synthesizing the concealed samples} and \textbf{2.} \emph{gradient projection}, which we discuss below.

\paragraph{Synthesizing the concealed samples.}
To obtain concealed samples that are visually dissimilar to sensitive data but whose gradient is similar to the sensitive data, we would like to solve the following optimization problem:
\begin{align}
    \label{eq: obj}
    &\min_{\tilde{\vx}_c} 1 - \frac{\big\langle \nabla_{\vtheta}\Ls(f_{\vtheta}(\tilde{\vx}_c), \tilde{\vy}_c) , 
    \nabla_{\vtheta}\Ls(f_{\vtheta}(\vx_{s}), \vy_{s}) \big\rangle 
    }{\left\| \nabla_{\vtheta}\Ls(f_{\vtheta}(\tilde{\vx}_c), \tilde{\vy}_c) \right\| \left\| \nabla_{\vtheta}\Ls(f_{\vtheta}(\vx_{s}), \vy_{s}) \right\|} \\
    \label{eq: obj2}
    &\max_{\tilde{\vx}_c} \left\| \tilde{\vx}_c - \vx_{s} \right\| \\
    \label{eq: obj3}
    & \hspace{1em} \text{s.t.} \big\| f_{\vtheta}(\tilde{\vx}_c) - f_{\vtheta}(\vx_{s}) \big\| \leq \epsilon.
\end{align}
We propose the following objective to achieve this
\begin{align}
\label{eq: f_obj}
    \Ls_{obj} &= (1 - \frac{\langle \nabla_{\vtheta}\Ls(f_{\vtheta}(\tilde{\vx}_c), \tilde{\vy}_c), \nabla_{\vtheta}\Ls(f_{\vtheta}(\vx_{s}), \vy_{s}) \rangle}{\left\| \nabla_{\vtheta}\Ls(f_{\vtheta}(\tilde{\vx}_c), \tilde{\vy}_c) \right\| \times \left\|\nabla_{\vtheta}\Ls(f_{\vtheta}(\vx_{s}), \vy_{s}) \right\|}) \notag \\
    & + e^{-\lambda_x \left\| \tilde{\vx}_c - \vx_{s} \right\|} + \lambda_z (\frac{\left\| f_{\vtheta}(\tilde{\vx}_c) - f_{\vtheta}(\vx_{s}) \right\|}{\left\|f_{\vtheta}(\vx_{s})\right\|} - \epsilon),
\end{align}
where $\lambda_z$ and $\lambda_x$ are hyperparameters to balance the different terms in the objective, $\epsilon$ controls the latent distance.
The first term and third term target achieving Goal-1 by ensuring that the concealed sample is similar to the sensitive data at the gradient level, while the second term learns the concealed sample to be visually dissimilar to the sensitive data.
\begin{remark}
The label corresponding to the concealed sample $\tilde{\vx}_c$ is denoted by $\tilde{\vy}_c$ in \cref{eq: f_obj}. To obtain $\tilde{\vx}_c$, we solve an optimization problem, starting from $\vx_0$, which may be a sample different from $\vx_s$. In such cases, we assign $\tilde{\vy}_c$ with the label of $\vx_0$, \ie, 
$\tilde{\vy}_c=\vy_0$. In our experiments, we show that $\tilde{\vx}_c$ can be randomly initialized, and accordingly, we set \(\tilde{\vy}_c\) at random. Our empirical evaluations in \textsection~\ref{sec: exp} show that the proposed method works equally well under both conditions.
\end{remark}

\paragraph{Gradient projection.}
Using \cref{eq: f_obj}, we can obtain the concealed sample $\vx_c$. What we need to do next is to ensure that the gradient of the mini-batch augmented with the concealed sample is aligned with the gradient of the original mini-batch, as this way, the server can improve its model. This will be achieved via the gradient projection, but before delving into details of projection and inspired by the mixup regularization~\cite{zhang2017mixup}, we propose an enhancement. Let $\vg$ be the gradient of the original mini-batch $\nabla_{\vtheta}\Ls(f_\vtheta(\vx_s), \vy_s)$. We obtain the gradient with the concealed sample as
\begin{align}
\label{eq: mixup}
    \vg_c \triangleq \nabla_{\vtheta} \Big\{ &\Ls(f_\vtheta(\vx_s), \vy_s) + \lambda_g \Ls(f_{\vtheta}(\tilde{\vx}_c), \tilde{\vy}_c) \notag \\
          &+ (1 - \lambda_g)\Ls(f_{\vtheta}(\tilde{\vx}_c), \vy_s) \Big\},
\end{align}
where $\lambda_g$ is a hyperparameter. Note that if $\lambda_g=1$, we indeed attain the gradient of the mini-batch augmented by the concealed sample.
However, including the gradient in the form $\nabla_{\vtheta}\Ls(f_{\vtheta}(\tilde{\vx}_c), \vy_s)$ is empirically observed to be beneficial. Analysis can be found in \textsection~\ref{sec: exp}.

To align the resulting gradient $\vg_c$ with the original gradient of the mini-batch $\vg$, we opt for the technique developed in \cite{lopez2017gradient}. This will ensure that the gradient sent to the server will improve the FL model. To this end, we compute the angle between the original gradient vector and the new gradient and check if it satisfies $\langle \vg, \vg_c \rangle \geq 0$.
If the constraints is satisfied, the new gradient $\vg_c$ behaves similarly to that of obtained from the mini-batch $\vx_s$; otherwise, we project the new gradient $\vg_c$ to the closest gradient $\hat{\vg}_{c}$ according to:
\begin{align}
\label{eq: proj}
    \argmin_{\hat{\vg}_{c}} \quad &\frac{1}{2} \left\| \vg_{c} - \hat{\vg}_{c} \right\|^{2}_{2}, \notag \\
    s.t. \quad &\langle \vg, \hat{\vg}_{c} \rangle \geq 0.
\end{align}
To efficiently solve \cref{eq: proj}, we employ the Quadratic Programming (QP) with inequality constraints:
\begin{align}
    \label{eq: proj_solu}
    \argmin_{v} \quad &\frac{1}{2} \vg^\top \vg v + \vg_{c}^\top \vg v, \notag \\
    \text{s.t.} \quad &v \geq 0.
\end{align}
The projected gradient $\hat{\vg}_{c}$ is given from the solution $v^*$ in \cref{eq: proj_solu} as $\hat{\vg}_{c}=\vg v^{*} + \vg_{c}$.
The complete pseudocode for the algorithm is provided in \cref{alg: pseudocode}.

\begin{algorithm}[tb]
\footnotesize
    \caption{\small{Defense by Concealing Sensitive Samples ($\text{DCS}^2$)}}
    \label{alg: pseudocode}
    \begin{algorithmic}[1]
    \Procedure{Gradient Obfuscation}{}
    \State $ \text{initialize the start point for constructing}$
    \Statex $ \hspace{5em} \text{the concealed data} \hspace{0.5em} \tilde{\vx}_c \gets \vx_0, \tilde{\vy}_c \gets \vy_0; $
    \State $ \text{get the concealed sample} \hspace{0.3em} \tilde{\vx}_c \gets \myEqref{eq: f_obj}; $
    \State $ \text{compute the new gradient} \hspace{0.3em} \vg_c \gets \myEqref{eq: mixup}; $
    \EndProcedure
    \Procedure{Gradient Projection}{}
    \State $ \text{get the gradient from the original batch}$
    \Statex $ \hspace{3em} \vg \gets \nabla_{\vtheta}\Ls(f_\vtheta(\vx_s), \vy_s); $
    \If {$\langle \vg, \vg_{c} \rangle < 0$}
    \State $ \text{get the solution} \hspace{0.3em} v^* \gets \myEqref{eq: proj_solu}; $
    \State $ \text{project the new gradient to the closest gradient}$
    \Statex$ \hspace{8em} \hat{\vg_c}=\vg v^{*} + \vg_{c}. $
    \EndIf
    \EndProcedure
    \end{algorithmic}
\end{algorithm}

\section{Experiments}
\label{sec: exp}
In this section, we first describe our evaluation settings, followed by a comparison of our defense with existing defenses against model inversion attacks in FL, to answer the following research questions (RQs):
\begin{description}[leftmargin=1.0cm,labelindent=0.1cm]
\small
\setlength\itemsep{0em}
    \item[RQ1:] Can the proposed method \(\text{DCS}^2\) effectively protect sensitive data against model inversion attacks in FL?
    \item[RQ2:] Is the proposed method \(\text{DCS}^2\) capable of maintaining FL performance while providing protection?
    \item[RQ3:] How does the proposed method \(\text{DCS}^2\) compare with existing defenses?
    \item[RQ4:] How does the proposed method \(\text{DCS}^2\) perform when defending against adaptive attacks?
    \item[RQ5:] How does the proposed method $\text{DCS}^2$ perform when the starting point for generating concealing samples varies?
\end{description}
Additional details, including the values of hyperparameters, are available in the supplementary material.

\subsection{Experimental setup}

\paragraph{Attack methods.}
We evaluate defenses against classical and state-of-the-art (SOTA) attacks in FL:
the improved version of the classical Deep Leakage from Gradients~\cite{zhu2019deep} called \textit{GS attack}~\cite{geiping2020inverting} that introduces image prior and uses cosine similarity as a distance metric to enhance reconstruction, and SOTA attack \textit{GGL attack} that uses a Generative Adversarial Network (GAN) to learn prior knowledge from public datasets.
We also include the recently proposed SOTA model modification attack \ie \textit{Imprint attack}~\cite{fowl2021robbing}.
Furthermore, we provide an evaluation when the \textit{adaptive attack} has strong prior knowledge about the private training data.

\paragraph{Defense baselines.}
Following recent works~\cite{sun2021soteria,gao2021privacy}, we compare our approach with defenses including \textit{DP-Gaussian} (adding Gaussian noise to gradients, following the implementation in~\citep{sun2021soteria,gao2021privacy}), and \textit{Prune} (Gradient Compression)~\cite{lin2017deep}.
We further compare against the recently proposed defense \textit{Soteria}~\cite{sun2021soteria}, which perturbs the representations.
In the supplementary material, we also provide a comparison with defenses that alter the sensitive data \eg, ATS~\cite{gao2021privacy}.

\paragraph{Datasets.}
We consider four datasets, namely MNIST~\cite{lecun1998gradient} with image resolution $28\times 28$, CIFAR10~\cite{krizhevsky2009learning} with image resolution $32\times 32$, CelebFaces Attributes (CelebA) Dataset~\cite{liu2015faceattributes} with image resolution rescaled to $32\times 32$ for a fair evaluation on GGL attack and TinyImageNet~\cite{le2015tiny} with image resolution rescaled to $224\times 224$.

\paragraph{Models.}
Being consistent with existing literature, we consider three model architectures i.e., LeNet \cite{lecun1998gradient} for MNIST, ConvNet (with the same structure as in Soteria~\cite{sun2021soteria}) for CIFAR10 and CelebA, ResNet18~\cite{he2016deep} for TinyImageNet.

\paragraph{Metrics.}
To quantify the quality of reconstructed images and compare them with the original sensitive data, we use peak signal-to-noise ratio (PSNR) as used in the work~\cite{balunovic2021bayesian}, and structural similarity index measure (SSIM)~\cite{wang2004image}. Besides, we use the learned perceptual image patch similarity (LPIPS) metric~\cite{zhang2018perceptual} for experiments on TinyImageNet.
When measuring PSNR and SSIM, lower values indicate better performances. When it comes to LPIPS, a higher number indicates a better performance.
We report classification accuracy values on the respective test sets and the protected data to measure the FL performance.

\paragraph{FL setting.}
We apply FedAvg~\cite{mcmahan2017communication} during training to report our results.
We set the local epoch $E$ as 1 (easier for attacks) and batch size $B$ as 64.
We have 10 clients in total; each client only has 200 samples on MNIST, 2000 samples on CIFAR10, 500 samples on CelebA, and 2000 samples on TinyImageNet.

\begin{table*}[tb]
    \centering
    \begin{adjustbox}{max width=0.99\textwidth}
    \begin{tabular}{lccccccccc}
        \toprule
        &\multicolumn{4}{c}{MNIST} & &\multicolumn{4}{c}{CIFAR10} \\
        \cmidrule{2-5} \cmidrule{7-10}
        Defense &PSNR$\downarrow$ &SSIM$\downarrow$ &Acc$\uparrow$ (Sensitive Data) &Acc$\uparrow$ (Test set) &
                &PSNR$\downarrow$ &SSIM$\downarrow$ &Acc$\uparrow$ (Sensitive Data) &Acc$\uparrow$ (Test set) \\
        \midrule
        None &59.20\scriptsize{$\pm$2.71} &1.00\scriptsize{$\pm$4.87} &86.98\scriptsize{$\pm$0.00} &87.16\scriptsize{$\pm$0.01} &
             &20.41\scriptsize{$\pm$3.15} &0.73\scriptsize{$\pm$0.09} &90.35\scriptsize{$\pm$0.04} &80.41\scriptsize{$\pm$0.01} \\
        \cdashlinelr{2-10}
        DP-Gaussian &35.38\scriptsize{$\pm$2.44} &0.83\scriptsize{$\pm$0.07} &85.94\scriptsize{$\pm$0.00} &86.91\scriptsize{$\pm$0.01} & &12.34\scriptsize{$\pm$1.34} &0.28\scriptsize{$\pm$0.06} &77.19\scriptsize{$\pm$0.18} &79.65\scriptsize{$\pm$0.04} \\
        Prune &14.13\scriptsize{$\pm$2.29} &0.37\scriptsize{$\pm$0.06} &85.94\scriptsize{$\pm$0.00} &86.91\scriptsize{$\pm$0.00} & 
              &11.26\scriptsize{$\pm$1.75} &0.22\scriptsize{$\pm$0.06} &77.80\scriptsize{$\pm$0.32} &79.51\scriptsize{$\pm$0.08} \\
        Soteria &9.67\scriptsize{$\pm$1.09} &0.30\scriptsize{$\pm$0.07} &86.98\scriptsize{$\pm$0.00} &86.94\scriptsize{$\pm$0.00} & 
                &11.48\scriptsize{$\pm$1.42} &0.29\scriptsize{$\pm$0.06} &\textbf{84.70\scriptsize{$\pm$0.32}} &79.76\scriptsize{$\pm$0.04} \\
        $\text{DCS}^2$ (Ours) &\textbf{7.84\scriptsize{$\pm$2.56}} &\textbf{0.17\scriptsize{$\pm$0.09}} &\textbf{86.98\scriptsize{$\pm$0.00}} &\textbf{86.98\scriptsize{$\pm$0.01}} & &\textbf{8.04\scriptsize{$\pm$1.10}} &\textbf{0.15\scriptsize{$\pm$0.05}} &80.39\scriptsize{$\pm$0.07} &\textbf{79.79\scriptsize{$\pm$0.03}} \\
        \bottomrule
    \end{tabular}
    \end{adjustbox}
    \vspace{-0.5em}
    \caption{Defenses against GS attack on MNIST and CIFAR10. Values are averaged. For DP-Gaussian, we follow the implementation in studies~\cite{sun2021soteria, gao2021privacy}.}
    \label{tab:mnist_cifar}
\end{table*}

\begin{table*}[tb]
    \centering
    \begin{adjustbox}{max width=0.99\textwidth}
    \begin{tabular}{lccccccccc}
        \toprule
        &\multicolumn{4}{c}{CelebA} & &\multicolumn{4}{c}{TinyImageNet} \\
        \cmidrule{2-5} \cmidrule{7-10}
        Defense &PSNR$\downarrow$ &SSIM$\downarrow$ &Acc$\uparrow$ (Sensitive Data) &Acc$\uparrow$ (Test set) &
                &SSIM$\downarrow$ &LPIPS$\uparrow$ &Acc$\uparrow$ (Sensitive Data) &Acc$\uparrow$ (Test set) \\
        \midrule
        None &19.92\scriptsize{$\pm$2.18} &0.75\scriptsize{$\pm$0.07} &100.0\scriptsize{$\pm$0.00} &93.79\scriptsize{$\pm$0.07} &
             &1.00\scriptsize{$\pm$0.00} &0.00\scriptsize{$\pm$0.00}  &73.94\scriptsize{$\pm$1.21} &66.41\scriptsize{$\pm$0.02} \\
        \cdashlinelr{2-10}
        DP-Gaussian &13.95\scriptsize{$\pm$1.52} &0.44\scriptsize{$\pm$0.08} &90.51\scriptsize{$\pm$0.47} &93.19\scriptsize{$\pm$0.04} &
                    &1.00\scriptsize{$\pm$0.00} &0.00\scriptsize{$\pm$0.00} &53.28\scriptsize{$\pm$0.78} &65.65\scriptsize{$\pm$0.07} \\
        Prune &9.57\scriptsize{$\pm$2.66} &0.24\scriptsize{$\pm$0.12} &91.41\scriptsize{$\pm$1.10} &93.25\scriptsize{$\pm$0.06} &
              &0.91\scriptsize{$\pm$0.12} &0.16\scriptsize{$\pm$0.20} &52.77\scriptsize{$\pm$0.07} &\textbf{65.73\scriptsize{$\pm$0.20}} \\
        Soteria &8.89\scriptsize{$\pm$2.63} &0.24\scriptsize{$\pm$0.11} &100.0\scriptsize{$\pm$0.00} &93.86\scriptsize{$\pm$0.01} &
                &1.00\scriptsize{$\pm$0.00} &0.00\scriptsize{$\pm$0.00} &41.84\scriptsize{$\pm$1.14} &52.06\scriptsize{$\pm$1.47} \\
        $\text{DCS}^2$ (Ours) &\textbf{8.24\scriptsize{$\pm$2.71}} &\textbf{0.17\scriptsize{$\pm$0.12}} &\textbf{100.0\scriptsize{$\pm$0.00}} &\textbf{94.31\scriptsize{$\pm$0.01}} & &\textbf{0.79\scriptsize{$\pm$0.22}} &\textbf{0.22\scriptsize{$\pm$0.23}} &\textbf{59.88\scriptsize{$\pm$0.71}} &65.68\scriptsize{$\pm$0.05} \\
        \bottomrule
    \end{tabular}
    \end{adjustbox}
    \vspace{-0.5em}
    \caption{Defenses against GGL attack on CelebA and Imprint attack on TinyImageNet. Values are averaged.}
    \label{tab:celeba_tinyimagenet}
    \vspace{-1em}
\end{table*}

\begin{table}[tb]
    \centering
    \begin{adjustbox}{max width=0.48\textwidth}
    \begin{tabular}{lcccc}
        \toprule
        $\lambda_g$ &SSIM$\downarrow$ &LPIPS$\uparrow$ &Acc$\uparrow$ (Sensitive Data) &Acc$\uparrow$ (Test set) \\ 
        \midrule
         0.5 &0.80\scriptsize{$\pm$0.20} &0.22\scriptsize{$\pm$0.21} &\textbf{60.33\scriptsize{$\pm$0.71}} &\textbf{65.76\scriptsize{$\pm$0.04}} \\
         0.7 &0.79\scriptsize{$\pm$0.22} &0.22\scriptsize{$\pm$0.23} &59.88\scriptsize{$\pm$0.71} &65.68\scriptsize{$\pm$0.05} \\
         1.0 &\textbf{0.78\scriptsize{$\pm$0.22}} &\textbf{0.23\scriptsize{$\pm$0.23}} &58.54\scriptsize{$\pm$0.46} &65.24\scriptsize{$\pm$0.21} \\
        \bottomrule
    \end{tabular}
    \end{adjustbox}
    \vspace{-0.5em}
    \caption{$\text{DCS}^2$ with different $\lambda_g$ on TinyImageNet.}
    \label{tab:lambda_g}
\end{table}

\subsection{Privacy-performance trade-off}
We consider 100\% of the training data in the target client as sensitive samples.
The optimal conditions for an adversary to invert gradients are a batch size of one, a low image resolution, and an untrained target network.

\paragraph{Results on MNIST and CIFAR10.}
We first evaluate defenses against the GS attack on the MNIST and CIFAR10 datasets using models with randomly initialized weights.
Results on \cref{tab:mnist_cifar} indicate that, compared with existing defenses, our proposed approach provides a better defense against the GS attack.
Specifically, on MNIST, the defense baselines reduce the PSNR from 59.20 to $\sim10$, while our defense can reduce the PSNR to around 8. On CIFAR10, our method reduces the SSIM to 0.17 when other defenses only reduce it to around 0.3.
In terms of the FL performance, as shown in \cref{tab:mnist_cifar}, our proposed defense method $\text{DCS}^2$ largely retains the performance compared with other defenses.
Specifically, on MNIST, when most defense baseline drops the performance by about 1\% on the sensitive data, our defense maintains the performance.

\paragraph{Results on CelebA and TinyImageNet.}
Further, we compare different defenses for more complex datasets, with larger capacity networks, on CelebA and TinyImageNet, to defend against stronger attacks.
We use randomly initialized weights and use the attribute gender as the target label in CelebA to perform binary classification. A pre-trained ResNet18 was applied for TinyImageNet.
As shown in Table~\ref{tab:celeba_tinyimagenet}, our defense provides the best protection while competitively maintaining the original FL performance.
Specifically, on CelebA, defending against the GGL attack, our method provides the best protection, and the FL performance is even improved while defenses DP-Gaussian and Prune drop by around 0.5\% on the test set.
On TinyImageNet, when defending against the Imprint attack, the defense Soteria cannot know where the adversary would insert the imprint module, so it cannot withstand the Imprint attack.
While most defenses cannot provide protection, our defense method increases the LPIPS from 0.00 to 0.22.

\cref{fig:dsc} shows the example of reconstructions from different attacks with defenses on different datasets.
The attacks could still recover some parts of the sensitive data with other defenses, while they fail with our proposed defense method.
The training process on various datasets with different defenses is illustrated in \cref{fig:loss_all}. Training with these defenses typically results in convergence. However, in the case of Soteria on TinyImageNet, approximately 90\% of the representations are perturbed, resulting in a convergence failure.

\cref{tab:lambda_g} presents the results for $\text{DCS}^2$ on TinyImageNet under varying values of $\lambda_g$.
As $\lambda_g$ increases, the protection for sensitive data improves. However, this leads to a reduction in the performance of the FL system.

\begin{figure}[tb]
    \centering
    \begin{tabular}[b]{ccc}
        \toprule
        Defense &PSNR$\downarrow$ &SSIM$\downarrow$ \\ 
        \midrule
        None                  &59.22\scriptsize{$\pm$2.71} &1.00\scriptsize{$\pm$4.77} \\
        $\text{DCS}^2$        &7.87\scriptsize{$\pm$2.44} &0.18\scriptsize{$\pm$0.09} \\
        \bottomrule
    \end{tabular}
    \includegraphics{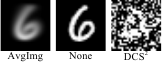}
    \captionlistentry[table]{Defend against adaptive attacks.}
    \label{tab:prior_attack}
    \captionsetup{labelformat=figtab}
    \caption{Defend against adaptive attacks.}
    \label{fig:prior_attack}
\end{figure}

\subsection{Comparison against adaptive attacks.}
We compared the proposed defense method $\text{DCS}^2$ against two SOTA attacks: Imprint and GGL. Imprint modifies the architecture, and GGL uses a GAN to learn prior knowledge from public datasets. 
As per ~\citet{gao2021privacy}, both these attacks are adaptive since the adversary ``starts the reconstruction from an image with certain semantic information''
or ``designs attack techniques instead of optimizing the distance between the real and dummy gradients''.
Results in \cref{tab:mnist_cifar,tab:celeba_tinyimagenet} indicate that our defense provides the best protection with minimal drop in accuracy.
For example, on TinyImageNet, our defense reduces the SSIM score from 1.0 to 0.79. In comparison, the defense Prune decreases it to approximately 0.9 and other defenses prove inadequate against this attack. The accuracy of the FL system using our defense on the sensitive data decreases by about 14\%, whereas other defenses drop exceeding 20\%.

Further, we design another strong attack where the adversary has strong prior knowledge and initializes the GS attack with the average image for each class.
Results are shown in \cref{tab:prior_attack}, our proposed method can still provide good protection against such an attack with prior knowledge about the sensitive data.
\cref{fig:prior_attack} shows an example of the reconstructions from this attack. The GS attack would initialize the dummy input with the AvgImg (average image) shown in \cref{fig:prior_attack}. The average image already explicitly includes information about the sensitive data, while our defense method could still protect the data against this adaptive attack.

\begin{figure}[tb]
  \centering
  \includegraphics[width=0.42\textwidth, keepaspectratio=True]{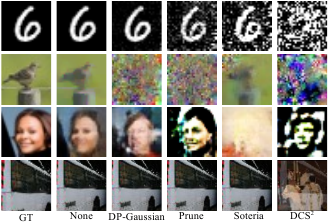}
  \vspace{-0.5em}
  \caption{Example of reconstructions for \cref{tab:mnist_cifar,tab:celeba_tinyimagenet}. From top to bottom are reconstructions from GS attacks on MNIST, those from GS attacks on CIFAR10, those from GGL attacks on CelebA, and those from Imprint attacks on TinyImageNet, respectively. (Best viewed in color)}
  \label{fig:dsc}
\end{figure}

\begin{table}[tb]
    \centering
    \begin{adjustbox}{max width=0.48\textwidth}
    \begin{tabular}{ccc|cccc}
        \hline
        MNIST &Noise &MixUP &PSNR$\downarrow$ &SSIM$\downarrow$ &Acc$\uparrow$ (Sensitive Data) &Acc$\uparrow$ (Test set) \\ 
        \hline
        \cmark &\xmark &\cmark &7.97\scriptsize{$\pm$2.48} &0.18\scriptsize{$\pm$0.08} &86.56\scriptsize{$\pm$0.57} &86.99\scriptsize{$\pm$0.01} \\
        \cmark &\xmark &\xmark &7.84\scriptsize{$\pm$2.56} &0.17\scriptsize{$\pm$0.09} &86.98\scriptsize{$\pm$0.00} &86.98\scriptsize{$\pm$0.01} \\
        \xmark &\cmark &\cmark &7.69\scriptsize{$\pm$2.38} &0.18\scriptsize{$\pm$0.08} &\textbf{86.98\scriptsize{$\pm$0.00}} &\textbf{86.99\scriptsize{$\pm$0.00}} \\
        \xmark &\cmark &\xmark &\textbf{7.40\scriptsize{$\pm$2.22}} &\textbf{0.16\scriptsize{$\pm$0.08}} &85.94\scriptsize{$\pm$0.00} &86.94\scriptsize{$\pm$0.00} \\
        \hline
    \end{tabular}
    \end{adjustbox}
    \vspace{-0.5em}
    \caption{Different start points on MNIST.}
    \label{tab:start_points}
    \vspace{-0.5em}
\end{table}

\begin{figure}[tb]
    \centering
    \begin{tabular}[b]{cccc}
        \toprule
        Defense &PSNR$\downarrow$ &SSIM$\downarrow$ &Acc$\uparrow$ \\ 
        \midrule
        None           &19.92\scriptsize{$\pm$2.18} &0.75\scriptsize{$\pm$0.07} &93.79\scriptsize{$\pm$0.07} \\
        $\text{DCS}^2$ &8.68\scriptsize{$\pm$2.78}  &0.18\scriptsize{$\pm$0.12} &94.13\scriptsize{$\pm$0.03} \\
        \bottomrule
    \end{tabular}
    \includegraphics[width=0.1\textwidth, keepaspectratio=True]{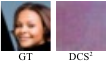}
    \captionlistentry[table]{Start points from CIFAR10 for CelebA.}
    \label{tab:st_celeba}
    \captionsetup{labelformat=figtab}
    \caption{Start points from CIFAR10 for CelebA.}
    \label{fig:st_celeba}
    \vspace{-0.5em}
\end{figure}

\subsection{Effect of Starting Points on Generating Concealed Samples}
We further evaluate our defense by choosing different initial starting points to craft the concealed samples.
\cref{tab:start_points} show the performance with different start points. `MixUP' means that $\tilde{\vx}_c$ is initialized with $0.7\vx_0+0.3\vx_s$.
\cref{tab:st_celeba} show the results when the start points are from CIFAR10, which has different distribution than the target task dataset CelebA.
As shown in \cref{tab:start_points,tab:st_celeba}, even starting from random noise and different domains, our defense method could still provide protection and retain the model's performance.

\subsection{Combination with existing defenses}
\begin{figure}[tb]
  \centering
  \includegraphics[width=0.48\textwidth, keepaspectratio=True]{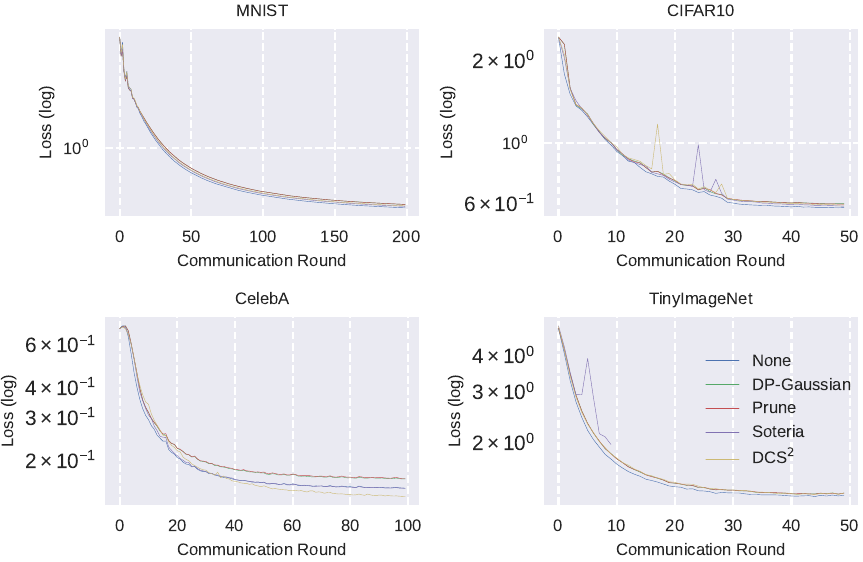}
  \vspace{-1em}
  \caption{Training process. (Best viewed in color)}
  \label{fig:loss_all}
\end{figure}

\begin{table}[tb]
    \centering
    \begin{adjustbox}{max width=0.48\textwidth}
    \begin{tabular}{lcccc}
        \toprule
        Defense &PSNR$\downarrow$ &SSIM$\downarrow$ &Acc$\uparrow$ (Sensitive Data) &Acc$\uparrow$ (Test set) \\ 
        \midrule
        Prune &14.13\scriptsize{$\pm$2.29} &0.37\scriptsize{$\pm$0.06} &85.94\scriptsize{$\pm$0.00} &86.91\scriptsize{$\pm$0.00} \\
        $\text{DCS}^2$ &7.84\scriptsize{$\pm$2.56} &0.17\scriptsize{$\pm$0.09} &\textbf{86.98\scriptsize{$\pm$0.00}} &\textbf{86.98\scriptsize{$\pm$0.01}} \\
        Prune$\&\text{DCS}^2$ &\textbf{6.08\scriptsize{$\pm$1.60}} &\textbf{0.12\scriptsize{$\pm$0.06}} &86.15\scriptsize{$\pm$0.47} &86.92\scriptsize{$\pm$0.01} \\
        \bottomrule
    \end{tabular}
    \end{adjustbox}
    \vspace{-0.5em}
    \caption{Combination of defenses.}
    \label{tab:combination}
    \vspace{-0.5em}
\end{table}

An illustration of combining $\text{DCS}^2$ with the defense `Prune' is presented in \cref{tab:combination}. In this scenario, the enhancement of protection for private training data is notable. While the performance experiences a slight decrease compared to the standalone proposed defense method, it still surpasses the performance of the defense `Prune' alone.

\section{Limitations}
\label{sec:dis}

While our empirical evaluations show that our proposed defense is effective in enhancing privacy and retaining FL performance, it requires additional computation to craft concealed samples (refer to the supplementary material for details on compute complexity).
Future directions to improve concealed samples-based defense include finding the best starting points and reducing the time to craft the concealed samples. We hope our defense can provide a new perspective for defending against model inversion attacks in FL.

\section{Conclusion}
In this work, we proposed an effective defense algorithm against model inversion attacks in FL. Our approach crafts concealed samples that imitate the sensitive data, but can obfuscate their gradients, thus making it challenging for an adversary to reconstruct sensitive data from the shared gradients.
To enhance the privacy of the sensitive data, the concealed samples are adaptively learned to be visually very dissimilar to the sensitive samples, while their gradients are aligned with the original samples to avoid FL performance drop. Our evaluations on four benchmark datasets showed that, compared with other defenses, our approach offers the best protection against model inversion attacks while simultaneously retaining or even improving the FL performance.

\section{Acknowledgments}
Mehrtash Harandi gratefully acknowledges the support from the Australian Research Council (ARC), project DP230101176.
\bibliography{aaai24}

\begin{thebibliography}{46}
\providecommand{\natexlab}[1]{#1}

\bibitem[{Abadi et~al.(2016)Abadi, Chu, Goodfellow, McMahan, Mironov, Talwar, and Zhang}]{abadi2016deep}
Abadi, M.; Chu, A.; Goodfellow, I.; McMahan, H.~B.; Mironov, I.; Talwar, K.; and Zhang, L. 2016.
\newblock Deep learning with differential privacy.
\newblock In \emph{Proceedings of the 2016 ACM SIGSAC conference on computer and communications security}, 308--318.

\bibitem[{Balunovi{\'c} et~al.(2022)Balunovi{\'c}, Dimitrov, Staab, and Vechev}]{balunovic2021bayesian}
Balunovi{\'c}, M.; Dimitrov, D.~I.; Staab, R.; and Vechev, M. 2022.
\newblock Bayesian Framework for Gradient Leakage.
\newblock \emph{ICLR}.

\bibitem[{Boenisch et~al.(2021)Boenisch, Dziedzic, Schuster, Shamsabadi, Shumailov, and Papernot}]{boenisch2021curious}
Boenisch, F.; Dziedzic, A.; Schuster, R.; Shamsabadi, A.~S.; Shumailov, I.; and Papernot, N. 2021.
\newblock When the Curious Abandon Honesty: Federated Learning Is Not Private.
\newblock \emph{arXiv preprint arXiv:2112.02918}.

\bibitem[{Bonawitz et~al.(2017)Bonawitz, Ivanov, Kreuter, Marcedone, McMahan, Patel, Ramage, Segal, and Seth}]{bonawitz2017practical}
Bonawitz, K.; Ivanov, V.; Kreuter, B.; Marcedone, A.; McMahan, H.~B.; Patel, S.; Ramage, D.; Segal, A.; and Seth, K. 2017.
\newblock Practical secure aggregation for privacy-preserving machine learning.
\newblock In \emph{proceedings of the 2017 ACM SIGSAC Conference on Computer and Communications Security}, 1175--1191.

\bibitem[{Carlini et~al.(2020)Carlini, Deng, Garg, Jha, Mahloujifar, Mahmoody, Song, Thakurta, and Tramer}]{carlini2020private}
Carlini, N.; Deng, S.; Garg, S.; Jha, S.; Mahloujifar, S.; Mahmoody, M.; Song, S.; Thakurta, A.; and Tramer, F. 2020.
\newblock Is Private Learning Possible with Instance Encoding?
\newblock \emph{arXiv preprint arXiv:2011.05315}.

\bibitem[{Dwork et~al.(2006)Dwork, McSherry, Nissim, and Smith}]{dwork2006calibrating}
Dwork, C.; McSherry, F.; Nissim, K.; and Smith, A. 2006.
\newblock Calibrating noise to sensitivity in private data analysis.
\newblock In \emph{Theory of cryptography conference}, 265--284. Springer.

\bibitem[{Fan et~al.(2020)Fan, Ng, Ju, Zhang, Liu, Chan, and Yang}]{fan2020rethinking}
Fan, L.; Ng, K.~W.; Ju, C.; Zhang, T.; Liu, C.; Chan, C.~S.; and Yang, Q. 2020.
\newblock Rethinking privacy preserving deep learning: How to evaluate and thwart privacy attacks.
\newblock In \emph{Federated Learning}, 32--50. Springer.

\bibitem[{Fowl et~al.(2022)Fowl, Geiping, Czaja, Goldblum, and Goldstein}]{fowl2021robbing}
Fowl, L.; Geiping, J.; Czaja, W.; Goldblum, M.; and Goldstein, T. 2022.
\newblock Robbing the Fed: Directly Obtaining Private Data in Federated Learning with Modified Models.
\newblock \emph{ICLR}.

\bibitem[{Gao et~al.(2021)Gao, Guo, Zhang, Qiu, Wen, and Liu}]{gao2021privacy}
Gao, W.; Guo, S.; Zhang, T.; Qiu, H.; Wen, Y.; and Liu, Y. 2021.
\newblock Privacy-preserving collaborative learning with automatic transformation search.
\newblock In \emph{Proceedings of the IEEE/CVF Conference on Computer Vision and Pattern Recognition}, 114--123.

\bibitem[{Geiping et~al.(2020)Geiping, Bauermeister, Dr{\"o}ge, and Moeller}]{geiping2020inverting}
Geiping, J.; Bauermeister, H.; Dr{\"o}ge, H.; and Moeller, M. 2020.
\newblock Inverting gradients-how easy is it to break privacy in federated learning?
\newblock \emph{Advances in Neural Information Processing Systems}, 33: 16937--16947.

\bibitem[{Goldreich(2009)}]{goldreich2009foundations}
Goldreich, O. 2009.
\newblock \emph{Foundations of cryptography: volume 2, basic applications}.
\newblock Cambridge university press.

\bibitem[{Griewank and Walther(2008)}]{griewank2008evaluating}
Griewank, A.; and Walther, A. 2008.
\newblock \emph{Evaluating derivatives: principles and techniques of algorithmic differentiation}.
\newblock SIAM.

\bibitem[{Gustafson et~al.(2023)Gustafson, Rolland, Ravi, Duval, Adcock, Fu, Hall, and Ross}]{gustafson2023facet}
Gustafson, L.; Rolland, C.; Ravi, N.; Duval, Q.; Adcock, A.; Fu, C.-Y.; Hall, M.; and Ross, C. 2023.
\newblock FACET: Fairness in Computer Vision Evaluation Benchmark.
\newblock In \emph{Proceedings of the IEEE/CVF International Conference on Computer Vision}, 20370--20382.

\bibitem[{He et~al.(2016)He, Zhang, Ren, and Sun}]{he2016deep}
He, K.; Zhang, X.; Ren, S.; and Sun, J. 2016.
\newblock Deep residual learning for image recognition.
\newblock In \emph{Proceedings of the IEEE conference on computer vision and pattern recognition}, 770--778.

\bibitem[{Huang et~al.(2021)Huang, Gupta, Song, Li, and Arora}]{huang2021evaluating}
Huang, Y.; Gupta, S.; Song, Z.; Li, K.; and Arora, S. 2021.
\newblock Evaluating gradient inversion attacks and defenses in federated learning.
\newblock \emph{Advances in Neural Information Processing Systems}, 34.

\bibitem[{Huang et~al.(2020)Huang, Song, Li, and Arora}]{huang2020instahide}
Huang, Y.; Song, Z.; Li, K.; and Arora, S. 2020.
\newblock Instahide: Instance-hiding schemes for private distributed learning.
\newblock In \emph{International Conference on Machine Learning}, 4507--4518. PMLR.

\bibitem[{Jeon et~al.(2021)Jeon, Lee, Oh, Ok et~al.}]{jeon2021gradient}
Jeon, J.; Lee, K.; Oh, S.; Ok, J.; et~al. 2021.
\newblock Gradient inversion with generative image prior.
\newblock \emph{Advances in Neural Information Processing Systems}, 34: 29898--29908.

\bibitem[{Jin et~al.(2021)Jin, Chen, Hsu, Yu, and Chen}]{jin2021catastrophic}
Jin, X.; Chen, P.-Y.; Hsu, C.-Y.; Yu, C.-M.; and Chen, T. 2021.
\newblock Catastrophic Data Leakage in Vertical Federated Learning.
\newblock \emph{Advances in Neural Information Processing Systems}, 34.

\bibitem[{Kairouz et~al.(2021)Kairouz, McMahan, Avent, Bellet, Bennis, Bhagoji, Bonawitz, Charles, Cormode, Cummings et~al.}]{kairouz2021advances}
Kairouz, P.; McMahan, H.~B.; Avent, B.; Bellet, A.; Bennis, M.; Bhagoji, A.~N.; Bonawitz, K.; Charles, Z.; Cormode, G.; Cummings, R.; et~al. 2021.
\newblock Advances and open problems in federated learning.
\newblock \emph{Foundations and Trends{\textregistered} in Machine Learning}, 14(1--2): 1--210.

\bibitem[{Krizhevsky, Hinton et~al.(2009)}]{krizhevsky2009learning}
Krizhevsky, A.; Hinton, G.; et~al. 2009.
\newblock Learning multiple layers of features from tiny images.

\bibitem[{Le and Yang(2015)}]{le2015tiny}
Le, Y.; and Yang, X. 2015.
\newblock Tiny imagenet visual recognition challenge.
\newblock \emph{CS 231N}, 7(7): 3.

\bibitem[{LeCun et~al.(1998)LeCun, Bottou, Bengio, and Haffner}]{lecun1998gradient}
LeCun, Y.; Bottou, L.; Bengio, Y.; and Haffner, P. 1998.
\newblock Gradient-based learning applied to document recognition.
\newblock \emph{Proceedings of the IEEE}, 86(11): 2278--2324.

\bibitem[{Lee et~al.(2021)Lee, Kim, Ahn, Hussain, Cho, and Son}]{lee2021digestive}
Lee, H.; Kim, J.; Ahn, S.; Hussain, R.; Cho, S.; and Son, J. 2021.
\newblock Digestive neural networks: A novel defense strategy against inference attacks in federated learning.
\newblock \emph{computers \& security}, 109: 102378.

\bibitem[{Li et~al.(2022)Li, Zhang, Liu, and Liu}]{li2022auditing}
Li, Z.; Zhang, J.; Liu, L.; and Liu, J. 2022.
\newblock Auditing Privacy Defenses in Federated Learning via Generative Gradient Leakage.
\newblock In \emph{Proceedings of the IEEE/CVF Conference on Computer Vision and Pattern Recognition}, 10132--10142.

\bibitem[{Lin et~al.(2017)Lin, Han, Mao, Wang, and Dally}]{lin2017deep}
Lin, Y.; Han, S.; Mao, H.; Wang, Y.; and Dally, W.~J. 2017.
\newblock Deep gradient compression: Reducing the communication bandwidth for distributed training.
\newblock \emph{arXiv preprint arXiv:1712.01887}.

\bibitem[{Liu et~al.(2015)Liu, Luo, Wang, and Tang}]{liu2015faceattributes}
Liu, Z.; Luo, P.; Wang, X.; and Tang, X. 2015.
\newblock Deep Learning Face Attributes in the Wild.
\newblock In \emph{Proceedings of International Conference on Computer Vision (ICCV)}.

\bibitem[{Lopez-Paz and Ranzato(2017)}]{lopez2017gradient}
Lopez-Paz, D.; and Ranzato, M. 2017.
\newblock Gradient episodic memory for continual learning.
\newblock \emph{Advances in neural information processing systems}, 30.

\bibitem[{McMahan et~al.(2017{\natexlab{a}})McMahan, Moore, Ramage, Hampson, and y~Arcas}]{mcmahan2017communication}
McMahan, B.; Moore, E.; Ramage, D.; Hampson, S.; and y~Arcas, B.~A. 2017{\natexlab{a}}.
\newblock Communication-efficient learning of deep networks from decentralized data.
\newblock In \emph{Artificial intelligence and statistics}, 1273--1282. PMLR.

\bibitem[{McMahan et~al.(2017{\natexlab{b}})McMahan, Ramage, Talwar, and Zhang}]{mcmahan2017learning}
McMahan, H.~B.; Ramage, D.; Talwar, K.; and Zhang, L. 2017{\natexlab{b}}.
\newblock Learning differentially private recurrent language models.
\newblock \emph{arXiv preprint arXiv:1710.06963}.

\bibitem[{Mo et~al.(2021)Mo, Borovykh, Malekzadeh, Haddadi, and Demetriou}]{mo2021quantifying}
Mo, F.; Borovykh, A.; Malekzadeh, M.; Haddadi, H.; and Demetriou, S. 2021.
\newblock Quantifying information leakage from gradients.
\newblock \emph{CoRR, abs/2105.13929}.

\bibitem[{Mohassel and Zhang(2017)}]{mohassel2017secureml}
Mohassel, P.; and Zhang, Y. 2017.
\newblock Secureml: A system for scalable privacy-preserving machine learning.
\newblock In \emph{2017 IEEE symposium on security and privacy (SP)}, 19--38. IEEE.

\bibitem[{Nguyen et~al.(2023)Nguyen, Chandrasegaran, Abdollahzadeh, and Cheung}]{nguyen2023re}
Nguyen, N.-B.; Chandrasegaran, K.; Abdollahzadeh, M.; and Cheung, N.-M. 2023.
\newblock Re-thinking Model Inversion Attacks Against Deep Neural Networks.
\newblock In \emph{Proceedings of the IEEE/CVF Conference on Computer Vision and Pattern Recognition}, 16384--16393.

\bibitem[{Scheliga, M{\"a}der, and Seeland(2022)}]{scheliga2022precode}
Scheliga, D.; M{\"a}der, P.; and Seeland, M. 2022.
\newblock PRECODE-A Generic Model Extension to Prevent Deep Gradient Leakage.
\newblock In \emph{Proceedings of the IEEE/CVF Winter Conference on Applications of Computer Vision}, 1849--1858.

\bibitem[{Song, Chaudhuri, and Sarwate(2013)}]{song2013stochastic}
Song, S.; Chaudhuri, K.; and Sarwate, A.~D. 2013.
\newblock Stochastic gradient descent with differentially private updates.
\newblock In \emph{2013 IEEE global conference on signal and information processing}, 245--248. IEEE.

\bibitem[{Sun et~al.(2021)Sun, Li, Wang, Yang, Li, and Chen}]{sun2021soteria}
Sun, J.; Li, A.; Wang, B.; Yang, H.; Li, H.; and Chen, Y. 2021.
\newblock Soteria: Provable defense against privacy leakage in federated learning from representation perspective.
\newblock In \emph{Proceedings of the IEEE/CVF Conference on Computer Vision and Pattern Recognition}, 9311--9319.

\bibitem[{Takahashi, Liu, and Liu(2023)}]{takahashi2023breaching}
Takahashi, H.; Liu, J.; and Liu, Y. 2023.
\newblock Breaching FedMD: Image Recovery via Paired-Logits Inversion Attack.
\newblock In \emph{Proceedings of the IEEE/CVF Conference on Computer Vision and Pattern Recognition}, 12198--12207.

\bibitem[{Voigt and Von~dem Bussche(2017)}]{voigt2017eu}
Voigt, P.; and Von~dem Bussche, A. 2017.
\newblock The eu general data protection regulation (gdpr).
\newblock \emph{A Practical Guide, 1st Ed., Cham: Springer International Publishing}, 10(3152676): 10--5555.

\bibitem[{Wang et~al.(2004)Wang, Bovik, Sheikh, and Simoncelli}]{wang2004image}
Wang, Z.; Bovik, A.~C.; Sheikh, H.~R.; and Simoncelli, E.~P. 2004.
\newblock Image quality assessment: from error visibility to structural similarity.
\newblock \emph{IEEE transactions on image processing}, 13(4): 600--612.

\bibitem[{Wei et~al.(2020)Wei, Liu, Loper, Chow, Gursoy, Truex, and Wu}]{wei2020framework}
Wei, W.; Liu, L.; Loper, M.; Chow, K.-H.; Gursoy, M.~E.; Truex, S.; and Wu, Y. 2020.
\newblock A framework for evaluating client privacy leakages in federated learning.
\newblock In \emph{European Symposium on Research in Computer Security}, 545--566. Springer.

\bibitem[{Wei et~al.(2021)Wei, Liu, Wut, Su, and Iyengar}]{wei2021gradient}
Wei, W.; Liu, L.; Wut, Y.; Su, G.; and Iyengar, A. 2021.
\newblock Gradient-leakage resilient federated learning.
\newblock In \emph{2021 IEEE 41st International Conference on Distributed Computing Systems (ICDCS)}, 797--807. IEEE.

\bibitem[{Yin et~al.(2021)Yin, Mallya, Vahdat, Alvarez, Kautz, and Molchanov}]{yin2021see}
Yin, H.; Mallya, A.; Vahdat, A.; Alvarez, J.~M.; Kautz, J.; and Molchanov, P. 2021.
\newblock See through gradients: Image batch recovery via gradinversion.
\newblock In \emph{Proceedings of the IEEE/CVF Conference on Computer Vision and Pattern Recognition}, 16337--16346.

\bibitem[{Zhang et~al.(2017)Zhang, Cisse, Dauphin, and Lopez-Paz}]{zhang2017mixup}
Zhang, H.; Cisse, M.; Dauphin, Y.~N.; and Lopez-Paz, D. 2017.
\newblock mixup: Beyond empirical risk minimization.
\newblock \emph{arXiv preprint arXiv:1710.09412}.

\bibitem[{Zhang et~al.(2018)Zhang, Isola, Efros, Shechtman, and Wang}]{zhang2018perceptual}
Zhang, R.; Isola, P.; Efros, A.~A.; Shechtman, E.; and Wang, O. 2018.
\newblock The Unreasonable Effectiveness of Deep Features as a Perceptual Metric.
\newblock In \emph{CVPR}.

\bibitem[{Zhao, Mopuri, and Bilen(2020)}]{zhao2020idlg}
Zhao, B.; Mopuri, K.~R.; and Bilen, H. 2020.
\newblock idlg: Improved deep leakage from gradients.
\newblock \emph{arXiv preprint arXiv:2001.02610}.

\bibitem[{Zhu and Blaschko(2020)}]{zhu2020r}
Zhu, J.; and Blaschko, M. 2020.
\newblock R-gap: Recursive gradient attack on privacy.
\newblock \emph{arXiv preprint arXiv:2010.07733}.

\bibitem[{Zhu, Liu, and Han(2019)}]{zhu2019deep}
Zhu, L.; Liu, Z.; and Han, S. 2019.
\newblock Deep leakage from gradients.
\newblock \emph{Advances in Neural Information Processing Systems}, 32.

\end{thebibliography}

\clearpage
\section{Appendix}
\label{sec:appendix}

\subsection{Compute Complexity.}
\label{sec:app_cc}
Let's consider a model with $d$ parameters, and we'll denote the time complexity for forward propagation as $h(d)$. As per the Baur-Strassen theorem~\cite{griewank2008evaluating}, the time complexity of a single step in backpropagation will be at most $5h(d)$.
Our approach introduces an additional time complexity of $6f(d)$ due to the incorporation of concealed samples.
Assuming that $\vx \in \mathbb{R}^{n}$ and $f(\vx)\in \mathbb{R}^{m}$, we can express the overall computational cost of the objective function in \cref{eq: f_obj} as $\gO(d^2+n^2+m^2)$.
If the perturbed gradient with concealed samples behaves dissimilar to that of the original gradient, then the computational cost of the gradient projection is around $\gO(d^3)$.
On TinyImageNet, we observed that one round of updates with our method on an NVIDIA RTX A100 GPU takes approximately 172 seconds. In comparison, an update without any defenses requires around 102 seconds.

\subsection{Model Architectures}
Details of the models used in this study are shown in \cref{tab: models_sup}. The activation layers of the model for the MNIST dataset are Sigmoid, and for CIFAR10 and CelebA, TinyImageNet datasets are ReLU.
\begin{table}[h]
  \centering
  \begin{adjustbox}{max width=0.45\textwidth}
  \begin{tabular}{ccc}
    \toprule
\rowcolor{lightblue!10}    MNIST  &CIFAR10/CelebA  &TinyImageNet \\
    \midrule
    $5 \times 5$ Conv, 12  &$5 \times 5$ Conv, 32
    &$7 \times 7$ Conv, 64\\
    $5 \times 5$ Conv, 12
    &$\begin{Bmatrix}5 \times 5 \text{ Conv, 64}\end{Bmatrix} \times 2$
    &$3 \times 3$ MaxPool\\
    $5 \times 5$ Conv, 12 
    &$\begin{Bmatrix}5 \times 5 \text{ Conv, 128}\end{Bmatrix} \times 3$
    &$\begin{Bmatrix}
        3 \times 3 \text{ Conv, 64}\\
        3 \times 3 \text{ Conv, 64}
      \end{Bmatrix} \times 2$\\
    $5 \times 5$ Conv, 12  
    &$3 \times 3$ MaxPool
    &$\begin{Bmatrix}
        3 \times 3 \text{ Conv, 128}\\
        3 \times 3 \text{ Conv, 128}
      \end{Bmatrix} \times 2$\\
    FC-10
    &$\begin{Bmatrix}
      5 \times 5 \text{ Conv, 128}
      \end{Bmatrix} \times 3$
    &$\begin{Bmatrix}
        3 \times 3 \text{ Conv, 256}\\
        3 \times 3 \text{ Conv, 256}
      \end{Bmatrix} \times 2$\\
    &$3 \times 3$ MaxPool
    &$\begin{Bmatrix}
        3 \times 3 \text{ Conv, 512}\\
        3 \times 3 \text{ Conv, 512}
      \end{Bmatrix} \times 2$\\
    &FC-10 (CIFAR10) / FC-2 (CelebA)
    &$7 \times 7$ AveragePool\\
    &
    &FC-200\\
    \bottomrule
  \end{tabular}
  \end{adjustbox}
  \caption{Model architectures for different datasets.}
  \label{tab: models_sup}
\end{table}

\subsection{Parameters and Details.}
\label{sec:app_param}
We build on the repository using the official implementation of the GS, GGL, and Imprint attack methods.
For the defenses Soteria, Prune, and DP-Gaussian, we build on the repository from the study~\cite{sun2021soteria}.
For the defense ATS, we build upon the repository from the study~\cite{balunovic2021bayesian}.
For training, we apply SGD optimizer and set the learning rate for updating the local models $\eta=0.01$ with exponential decay.
The pruning rate from the defense Prune is 0.9 for CIFAR10 and 0.7 for others.
The variance of noise distribution from the defense DP-Gaussian is 0.01 for MNIST, 0.5 for TinyImageNet, and 0.001 for others.
The pruning rate from the defense Soteria is 0.2 for MNIST, 0.5 for CIFAR10, 0.7 for CelebA, and 0.9 for TinyImageNet.
Our method set $\epsilon=0.01$, $\lambda_g=0.7$, and the number of iteration as 1000 for all datasets, $\lambda_x=0.01$ and $\lambda_z=0.01$ for MNIST and CIFAR10, $\lambda_x=0.01$ and $\lambda_z=0.1$ for CelebA, $\lambda_x=0.001$ and $\lambda_z=0.01$ for TinyImageNet.
The weights of penalty terms in Eq. (9) are chosen to balance them or kept fixed across all experiments (\eg, $\epsilon = 0.01$ for all experiments.)
In our experiments, we opt for the entirety of the training data in the target client to be sensitive and generate one synthetic sample per sensitive datum (ratio 1:1).
SSIM is 0.92 with an accuracy of 87\% and 0.86 with an accuracy of 86.91\% for ratios 10:1 and 5:1, respectively.
While the protection becomes better, the model's performance will drop.

\begin{figure}[H]
  \centering
  \includegraphics[width=0.48\textwidth, keepaspectratio=True]{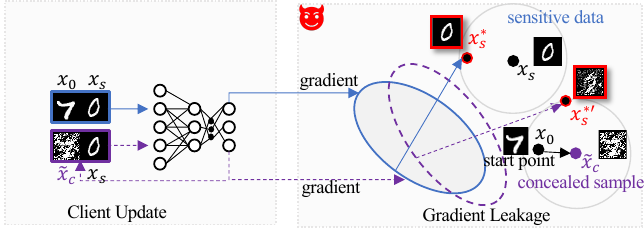}
  \caption{Illustration of gradient leakage and the proposed defense method. The adversary can obtain a perfect reconstruction $\vx_{s}^{*}$ for the sensitive data $\vx_s$ when the gradient is over the input $\vx_0$ and $\vx_s$, but it fails and get the reconstruction $\vx_{s}^{*'}$ when the gradient is over our concealed sample $\tilde{\vx}_c$ and $\vx_s$.}
  \label{fig: alg}
\end{figure}

Consider a sensitive data point $\vx_s$ (see \cref{fig: alg} for an illustration), we aim to craft the concealed sample $\tilde{\vx}_c$ which makes $\nabla_{\vtheta}\Ls(f_{\vtheta}(\tilde{\vx}_c), \tilde{\vy}_c)$ approaching $\nabla_{\vtheta}\Ls(f_{\vtheta}(\vx_s), \vy_s)$.
This strategy obfuscates the sensitive samples, as the reconstruction by the adversary through the gradient will contain information from the concealed sample, which we aim to be visually different from the sensitive data.

\subsection{Extra Results}

\begin{table}[H]
    \centering
    \begin{adjustbox}{max width=0.48\textwidth}
    \begin{tabular}{lcccc}
        \toprule
        Defense &PSNR$\downarrow$ &SSIM$\downarrow$ &Acc$\uparrow$ (Sensitive Data) &Acc$\uparrow$ (Test set) \\ 
        \midrule
        ATS &19.68\scriptsize{$\pm$4.60} &0.59\scriptsize{$\pm$0.14} &67.99\scriptsize{$\pm$0.57} &75.70\scriptsize{$\pm$0.02} \\ 
        \rowcolor{grey!10} $\text{DCS}^2$ (Ours) &\textbf{8.04\scriptsize{$\pm$1.10} }&\textbf{0.15\scriptsize{$\pm$0.05}}  &\textbf{80.39\scriptsize{$\pm$0.07}} &\textbf{79.79\scriptsize{$\pm$0.03}} \\
        \bottomrule
    \end{tabular}
    \end{adjustbox}
    \caption{Defenses against GS attack on CIFAR10. ATS applies data augmentations and here it needs to run for extra 50 rounds to converge.}
    \label{tab:ats}
\end{table}

\begin{table}[H]
    \centering
    \begin{adjustbox}{max width=0.48\textwidth}
    \begin{tabular}{lcccc}
        \toprule
        Defense &PSNR$\downarrow$ &SSIM$\downarrow$ &Acc$\uparrow$ (Sensitive Data) &Acc$\uparrow$ (Test set) \\ 
        \midrule
        None &57.50\scriptsize{$\pm$1.95} &1.00\scriptsize{$\pm$0.00} &89.84\scriptsize{$\pm$0.00} &76.60\scriptsize{$\pm$0.00} \\
        DP-Gaussian &34.73\scriptsize{$\pm$0.79} &0.83\scriptsize{$\pm$0.04} &83.59\scriptsize{$\pm$0.00} &75.75\scriptsize{$\pm$0.01} \\
        Prune &14.49\scriptsize{$\pm$1.81} &0.39\scriptsize{$\pm$0.05} &83.59\scriptsize{$\pm$0.00} &75.75\scriptsize{$\pm$0.00} \\
        Soteria &7.28\scriptsize{$\pm$0.60} &0.23\scriptsize{$\pm$0.04} &85.42\scriptsize{$\pm$0.00} &\textbf{75.92\scriptsize{$\pm$0.01}} \\
        \rowcolor{grey!10} $\text{DCS}^2$ (Ours) &\textbf{7.27\scriptsize{$\pm$1.77}} &\textbf{0.16\scriptsize{$\pm$0.06}} &\textbf{89.32\scriptsize{$\pm$0.00}} &75.19\scriptsize{$\pm$0.02} \\
        \bottomrule
    \end{tabular}
    \end{adjustbox}
    \caption{Against GS attack on MNIST on Non-IID setting. Non-IID data indeed presents additional challenges to training.}
    \label{tab:noniid}
\end{table}

\begin{figure}[H]
  \centering
  \includegraphics[width=0.48\textwidth, keepaspectratio=True]{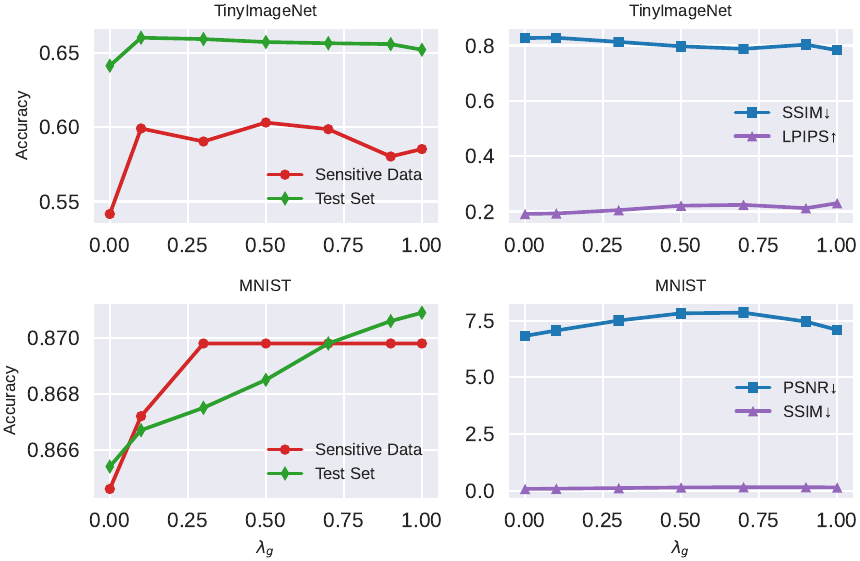}
  \caption{$\text{DCS}^2$ with different $\lambda_g$.}
  \label{fig:lambda_g}
\end{figure}

\begin{figure}[H]
  \centering
  \includegraphics[width=0.48\textwidth, keepaspectratio=True]{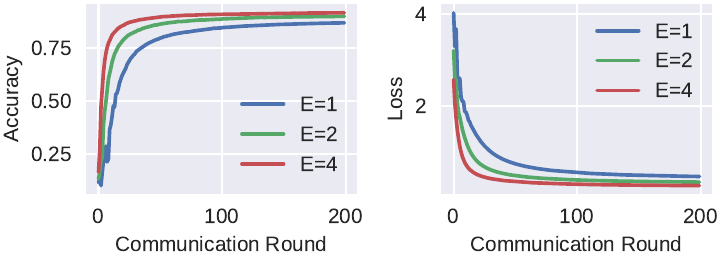}
  \caption{FL training process of $\text{DCS}^2$ on MNIST with different number of local epochs.}
  
  \label{fig:localE}
\end{figure}

\begin{table}[H]
\scriptsize
\centering
\label{tab:prior_attack1}
\begin{tabular}{l|ccccc}
    \hline
    &None &DP-Gaussian &Prune &Soteria &$\text{DCS}^2$ \\
    \hline
    PSNR$\downarrow$ &59.22\scriptsize{$\pm$2.71} &35.28\scriptsize{$\pm$2.50} &14.23\scriptsize{$\pm$2.23} &9.94\scriptsize{$\pm$1.10} &\textbf{7.87\scriptsize{$\pm$2.44}} \\
    SSIM$\downarrow$ &1.00\scriptsize{$\pm$4.77}  &0.82\scriptsize{$\pm$0.07} &0.37\scriptsize{$\pm$0.06} &0.31\scriptsize{$\pm$0.07} &\textbf{0.18\scriptsize{$\pm$0.09}} \\
    \hline
\end{tabular}
\caption{Defend against adaptive attacks on MNIST.}
\end{table}

\begin{figure}[H]
  \centering
  \includegraphics[width=0.48\textwidth, keepaspectratio=True]{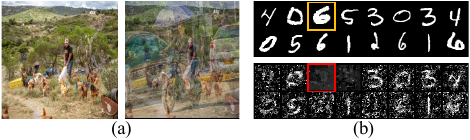}
  \caption{(a) Left to Right: reconstructions by the Imprint attack \without and \with $\text{DCS}^2$ on FACET. (b) Top to Bottom: GT and reconstructions by the GS attack on MNIST.}
  \label{fig:facet_and_bs16}
\end{figure}

We evaluated the Imprint attack with the recently introduced dataset FACET~\cite{gustafson2023facet} (\cref{fig:facet_and_bs16} (a)). With our defense applied, the attack failed to perfectly reconstruct the data. The accuracy of the model on the test set \without and \with our defense is around 77.21\% and 76.62\%, respectively.

Note that neither optimization-based attacks nor model modification attacks can precisely separate the gradient for individual data points.
In \cref{fig:facet_and_bs16} (b), for instance, there are four `6' within the batch; the gradients \wrt these images cannot be fully separated, and the GS attack fails to reconstruct the third image.
Besides, similar to any defense mechanism, it is possible to identify the concealed samples.
For example, an adversary could potentially modify the model to reconstruct the concealed samples and use a filtering mechanism to identify them.
Our model might become vulnerable if the adversary has extensive knowledge about the data and model as exemplified by the GGL attack (use partial training data to learn prior knowledge), or has the ability to modify model architecture as in the Imprint attack. Fig.2 and \cref{fig:facet_and_bs16} (a) provide examples of GGL and Imprint attacks, where the attack managed to reconstruct facial outlines and vague information. It is important to note that model modification methods are generally identifiable and can be countered by vigilant clients.

\end{document}